\documentclass[letterpaper, 10pt, conference]{ieeeconf}      

\usepackage{times}
\usepackage{epsfig}
\usepackage{graphicx}
\usepackage{amsmath}
\usepackage{amssymb}
\usepackage{textcomp}
\usepackage{multirow}
\usepackage{adjustbox}

\usepackage{multirow}
\usepackage{colortbl}
\usepackage{hhline}
\usepackage{subcaption}
\usepackage{verbatim}

\usepackage{gensymb}
\usepackage{xcolor}

\makeatletter
\let\NAT@parse\undefined
\DeclareRobustCommand\onedot{\futurelet\@let@token\@onedot}
\def\@onedot{\ifx\@let@token.\else.\null\fi\xspace}

\makeatother

\usepackage[colorlinks,pagebackref=false,citecolor=blue,bookmarks=false,hypertexnames=true]{hyperref}

\IEEEoverridecommandlockouts                              

\title{\LARGE \bf
VM-MODNet: Vehicle Motion aware Moving Object Detection \\ for Autonomous Driving}

\author{Hazem Rashed$^{1}$, Ahmad El Sallab$^{1}$ and Senthil Yogamani$^{2}$ \\ %
$^{1}$Valeo R\&D Cairo, Egypt \quad\quad $^{2}$Valeo Visions Systems, Ireland \\
         {\tt \small\{hazem.rashed, ahmad.el-sallab, senthil.yogamani\}@valeo.com}
}

\begin{document}

\maketitle
\thispagestyle{empty}
\pagestyle{empty}

\begin{abstract}
Moving object Detection (MOD) is a critical task in autonomous driving as moving agents around the ego-vehicle need to be accurately detected for safe trajectory planning. It also enables appearance agnostic detection of objects based on motion cues. There are geometric challenges like motion-parallax ambiguity which makes it a difficult problem.  
In this work, we aim to leverage the vehicle motion information and feed it into the model to have an adaptation mechanism based on ego-motion. The motivation is to enable the model to implicitly perform ego-motion compensation to improve performance. We convert the six degrees of freedom vehicle motion into a pixel-wise tensor which can be fed as input to the CNN model. 
The proposed model using Vehicle Motion Tensor (VMT) achieves an absolute improvement of 5.6\% in mIoU over the baseline architecture. We also achieve state-of-the-art results on the public KITTI\_MoSeg\_Extended dataset even compared to methods which make use of LiDAR and additional input frames. Our model is also lightweight and runs at 85 fps on a TitanX GPU. Qualitative results are provided in \url{https://youtu.be/ezbfjti-kTk}.
\end{abstract}

\section{INTRODUCTION}

Autonomous Driving (AD) environment is complex as they include moving objects that are navigating in different ways \cite{horgan2015vision, eising2021near}. 
Thus, full perception of all the surround moving agents is necessary for effective motion planning of the autonomous vehicle. Motion in the image, captured by optical flow, is induced by motion of other objects in the scene and by the ego-motion of the vehicle carrying the reference camera.  
Unlike systems where the reference camera is fixed, it is challenging to
predict moving objects in automotive scenes as the ego-vehicle is constantly in motion. 

Motion is a strong cue in automotive scenes. Moving objects like pedestrians and vehicles pose a higher risk and it is essential to reliably detect them. Motion cues also be used to detect objects in an appearance agnostic manner. For example,  construction trucks and animals like moose which are rare to be trained based on appearance cues can be alternatively be detected using motion cues. In addition, High Definition maps which is a main source for autonomous driving provides a reliable prior detection of static objects \cite{ravi2018real}. Thus, it is more important to detect moving objects reliably. 

Dominant approaches for moving object detection (MOD) make use of a combination of optical flow and RGB images to combine motion and appearance cues. They are either combined in early or mid level fusion techniques within a CNN model. These methods do not take advantage of known vehicle motion. Vehicle motion has six degrees of freedom comprising of three rotation angles and three translations. It can be obtained by a highly accurate inertial measurement unit (IMU) sensor. It can also be partially obtained by vehicle odometry sensors which provides two translations $(dx, dy)$ and yaw angle from steering wheel. It can also be estimated using visual odometry or from a sensor fused odometry.


\begin{figure}[!t]
\centering
\includegraphics[width=\columnwidth]{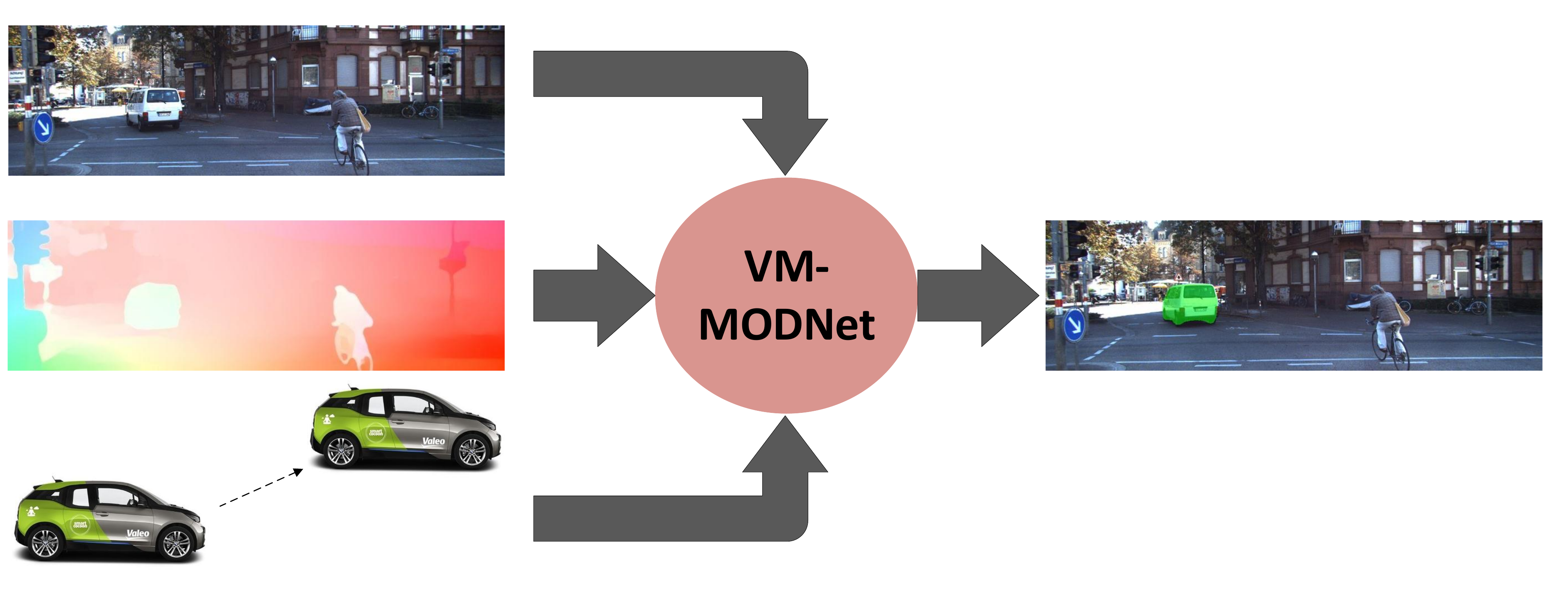}
\caption{Our model predicts motion segmentation using RGB image, optical flow and vehicle motion as inputs.} 
\vspace{-0.51cm}
\label{fig:CartesianPC}
\end{figure}

In this work, we aim to leverage vehicle motion and use it explicitly within a CNN model as inductive bias to improve the accuracy. We propose and implement a
vehicle motion aware network that will be able to detect surrounding moving objects more accurately. In this work, we focus on KITTI \cite{Geiger2012CVPR} dataset which provides accurate IMU values for vehicle motion. This work can also be used for other estimates of vehicle motion mentioned in the previous paragraph. We convert vehicle motion parameters into a pixel-wise vehicle motion tensor (VMT) that is suitable for using within a CNN network. We demonstrate significant improvements over the baseline using VMT.
To summarize, the contributions of this work include:

\begin{itemize}
    \item Design and implementation of CNN based moving object detection utilizing vehicle motion.
    \item State-of-the-art results on KITTI\_MoSeg\_Extended dataset with significantly faster runtime.
    \item Ablation study of different fusion methodologies.
\end{itemize}

The paper is organized as follows. Section \ref{sec:relatedwork} reviews the related work in moving object detection for autonomous driving. Section \ref{sec:methodology} discusses the proposed architecture including details of modelling of vehicle motion tensor. Section \ref{sec:experiments} describes the experimental setup and discusses quantitative and qualitative analysis.  Finally, Section \ref{sec:conc} provides concluding remarks.

\section{RELATED WORK} \label{sec:relatedwork}

Motion detection has been studied through classical approaches such as \cite{menze2015object}. Recently, CNN-based approaches provide better accuracy as they can encode global context. However, they require a extensive dataset with diverse moving objects. 
Optical flow has been used in generic foreground segmentation by \cite{jain2017fusionseg}. In \cite{drayer2016object,tokmakov2017learning}, video object segmentation has been explored, however these models are computationally intensive.
Siam et al. \cite{siam2018modnet} explored motion segmentation using CNN for autonomous driving scenario and further improved it by using depth \cite{siam2018real}.  
In \cite{yahiaoui2019fisheyemodnet}, MOD has been explored on images captured by wide-angle fisheye camera using \cite{yogamani2019woodscape} dataset.

In addition to camera sensors, MOD has been explored using LiDAR sensors as well. The most common way is to predict moving points using geometric constraints and then perform clustering to obtain moving objects \cite {dewan2016motion}. Using deep learning approaches, 3D convolution has been utilized to predict moving vehicles in \cite {li20173d}. 
In other approaches, the points were projected from 3D to 2D range images to make use of conventional 2D convolutions \cite {li2016vehicle}. 
In FuseMODNet \cite{rashed2019fusemodnet}, optical flow was generated from both camera and LiDAR to obtain robust low-illumination detection. Some approaches predict MOD using two sequential LiDAR scans to implicitly learn motion without using optical flow \cite{dewan2017deep}.


\section{PROPOSED METHOD} \label{sec:methodology}



In this section, we describe the proposed method including vehicle motion modelling and our network architecture.

\subsection{Baseline Architecture}
We start with the OmniDet \cite{ravikumar2021omnidet} motion segmentation network using  two-stream RGB only network. The network consists of two ResNet18 streams with shared weights and a motion segmentation decoder with deconv layers for upsampling to the higher resolution output. The architecture is simple and it provides real-time performance.
We train this model by feeding an RGB image at time $t$ in one stream, and the previous image at time $t-1$ in another stream. This model is reported as the baseline  in the first row of Table \ref{tab:results}. To enable comparison of accuracy and inference speeds with previous methods \cite{rashed2019fusemodnet,ramzy2019rst}, we scale the input resolution to $(1224,256)$. Then we replace the previous frame with Optical flow to enable better fusion with Vehicle Motion Tensor. This two-stream RGB and Optical flow model with shared weights will act as our baseline architecture.

\subsection{Vehicle Motion Tensor}

\textbf{Motivation:}
Motion segmentation is far more challenging in autonomous vehicles compared to surveillance or traffic light cameras which are fixed. The camera motion due to motion of the vehicle induces optical flow in all the static objects and it becomes difficult to separate them from moving objects. In addition, there are fundamental geometric challenges like motion-parallax ambiguity which makes it difficult to distinguish between a parallel moving car versus a static car. 

In this work, we aim to design a vehicle motion aware network that utilizes ego-motion information to improve motion segmentation.
Ego-motion compensation is commonly used in classical computer vision to subtract the camera motion related motion fields in optical flow. It was explored for motion segmentation task in SMSNet \cite{vertens2017smsnet}, where optical flow was compensated utilizing depth information. It requires a good depth estimation model which is a complex task on its own and does not handle noise in ego-motion as the compensation term is explicitly subtracted from the optical flow. In contrast, we want to use even noisy ego-motion from vehicle odometry sensors without using depth. 

Camera motion and vehicle motion are equivalent except for a possible co-ordinate system transformation. 
Camera motion information has six degrees of freedom. It is difficult to incorporate these six scalar values directly as input to a CNN model. Thus, we explore to convert them to a pixel-wise tensor which is easy to fuse with other image planes. We make use of the concept of Motion Fields which is closely related to Optical Flow. Motion field is the 2D vector field of velocities of image pixels induced by relative motion between observing camera and the 3D scene. Motion field is the projection of 3D relative velocity vectors onto image plane whereas optical flow is the observed 2D displacements on the image plane without using 3D information. We propose to use a modified version of motion fields as a pixel-wise encoding of ego-motion in a scene agnostic manner. Then we feed this as an independent input plane so that the network can learn to combine it effectively with optical flow and image information. This design will enable the network to have a loose coupling on ego-motion so that the estimation does not break down when ego-motion is not correct unlike an explicit ego-motion compensation.


We provide a short overview of motion fields in this section, please refer to this textbook \cite{trucco1998introductory} for more details. We make use of a pinhole camera model for the KITTI images. They have a slight barrel distortion which can be rectified using the intrinsic parameters. Motion fields $\vec{v}$ induced by the camera motion has two components for rotation $\vec{v_r}$ and translation $\vec{v_t}$.
\begin{align}
\vec{v}
= \vec{v_r} + \vec{v_t}
\end{align}



\begin{figure*}[t]
\captionsetup[subfigure]{labelformat=empty}
\centering
\begin{adjustbox}{minipage=\linewidth,scale=0.8}
\begin{subfigure}{.5\textwidth}
    \includegraphics[width=\columnwidth]{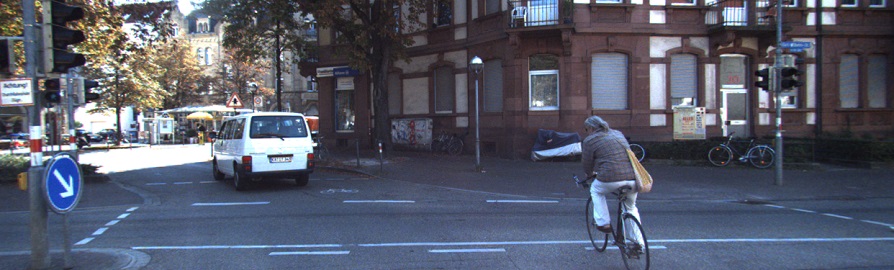}
    \vspace{-1cm}
    \caption{\textcolor{white}{(a)}}
\end{subfigure}%
\hfill
\begin{subfigure}{.5\textwidth}
    \includegraphics[width=\columnwidth]{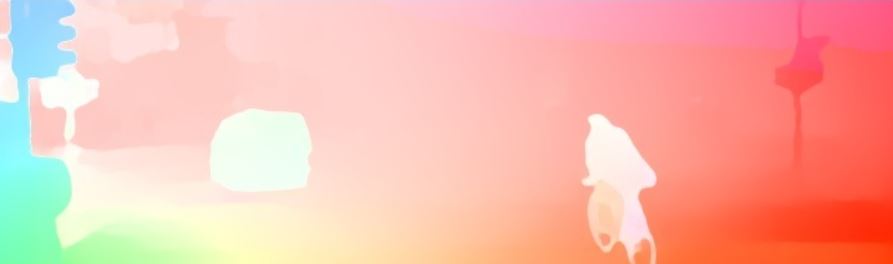}
    \vspace{-1cm}
    \caption{\textcolor{black}{(e)}}
\end{subfigure}%

\begin{subfigure}{.5\textwidth}
    \includegraphics[width=\columnwidth]{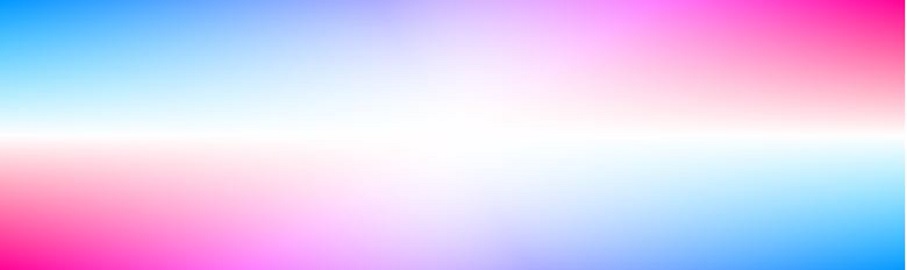}\\
    \vspace{-1cm}
    \caption{\textcolor{black}{(b)}}
\end{subfigure}%
\hfill
\begin{subfigure}{.5\textwidth}
    \includegraphics[width=\columnwidth]{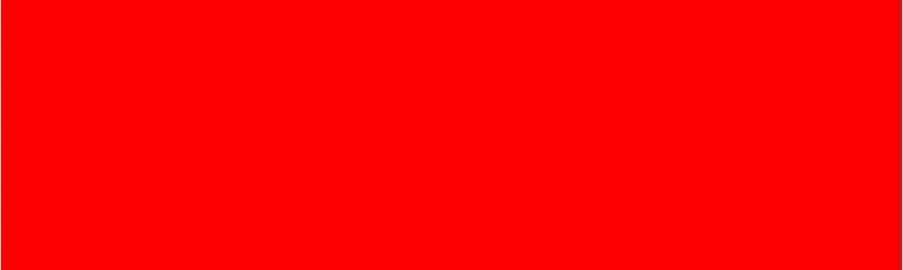}
    \vspace{-1cm}
    \caption{\textcolor{black}{(f)}}
\end{subfigure}%

\begin{subfigure}{.5\textwidth}
    \includegraphics[width=\columnwidth]{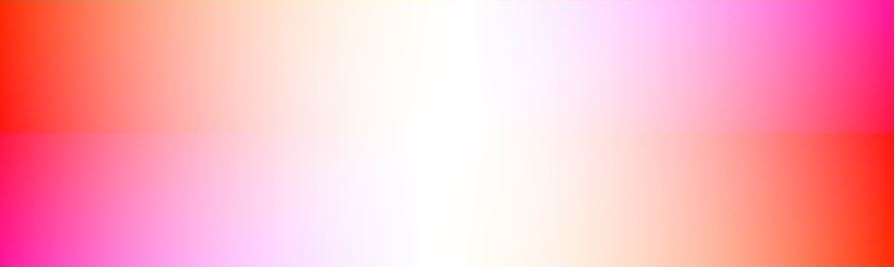}
    \vspace{-1cm}
    \caption{\textcolor{black}{(c)}}
\end{subfigure}%
\hfill
\begin{subfigure}{.5\textwidth}
    \includegraphics[width=\columnwidth]{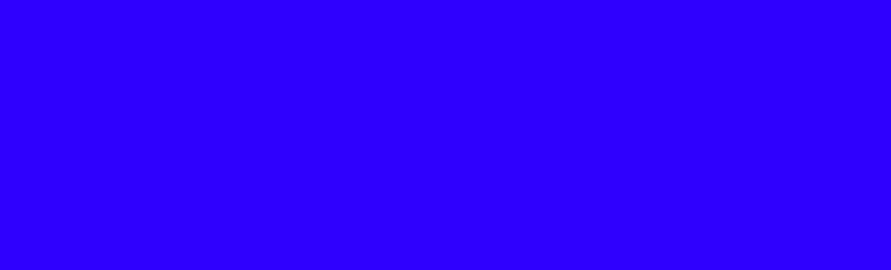}
    \vspace{-1cm}
    \caption{\textcolor{black}{(g)}}
\end{subfigure}%

\begin{subfigure}{.5\textwidth}
    \includegraphics[width=\columnwidth]{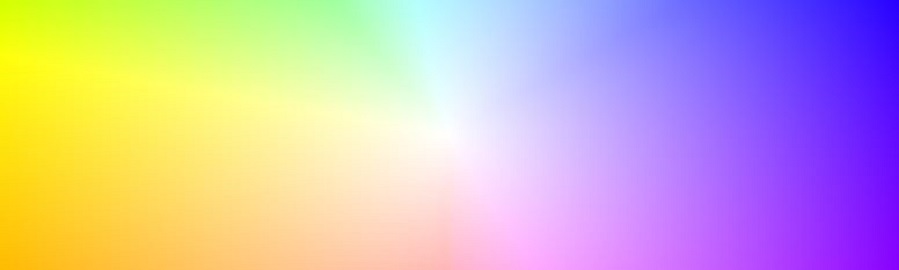}
    \vspace{-1cm}
    \caption{\textcolor{black}{(d)}}
\end{subfigure}%
\hfill
\begin{subfigure}{.5\textwidth}
    \includegraphics[width=\columnwidth]{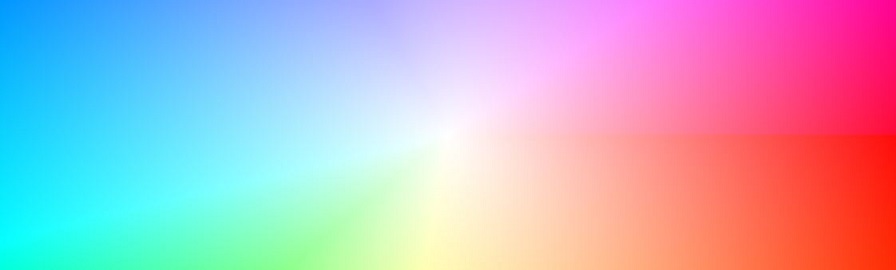}
    \vspace{-1cm}
    \caption{\textcolor{black}{(h)}}
\end{subfigure}%
\quad
\end{adjustbox}
    \caption{Different representations of Vehicle Motion Tensor (VMT) computed from six degrees of freedom vehicle motion. \textbf{(a)} is the RGB image and \textbf{(e)} is optical flow map. \textbf{(b,c,d)} correspond to VMT planes generated from each camera rotation angle around $(x,y,z)$ axes. \textbf{(f,g,h)} correspond to VMT planes generated from each camera translation in directions of $(x,y,z)$ axes. All the images are normalized for better visualization.}
    \label{fig:motion_map_representations}
    \vspace{-0.4cm}
  
\end{figure*}

\subsubsection{Rotation}
Rotation of the camera can be described by three parameters namely pitch, roll and yaw. Motion fields induced by pure camera rotation is derived in  \cite{trucco1998introductory} as shown below. It is important to observe that it is independent of scene structure.
\begin{align}
\vec{v_r}
=
\begin{pmatrix}
u_r \\
v_r
\end{pmatrix}
=
\begin{pmatrix}
- \omega_yf  + \omega_zy + \dfrac{\omega_x}{f}xy  - \dfrac{\omega_y}{f}x^{2} \\
\omega_xf - \omega_zx - \dfrac{\omega_y}{f}xy + \dfrac{\omega_x}{f}y^{2}  
\end{pmatrix} 
\end{align}

where $(\omega_x,\omega_y,\omega_z)$ represent the camera rotation parameters, $(x,y)$ represent the pixel indices and $(u_r,v_r)$ represent motion vectors corresponding to camera motion in the image coordinate system. The visualization of this equation is illustrated in the first column of Fig. \ref{fig:motion_map_representations}. Rows (b) ,(c) and (d) represents rotation around x, y and z axis respectively. These motion vectors are converted to colorwheel representation to enable reuse of CNN pre-trained weights similar to how it was done for optical flow in \cite{siam2018modnet}. 
To compute each component separately, we assume that the other components are equal to zero. 
Focal length is normalized to 1 and the output vectors are normalized from 0 to 255 for better visualization. The blue color represents pixels moving to the left direction which indicates the camera rotation in the right direction, while the red color indicates the opposite.

\begin{figure}[!t]
\centering
\begin{adjustbox}{minipage=\linewidth,scale=0.7}
\includegraphics[width=\columnwidth]{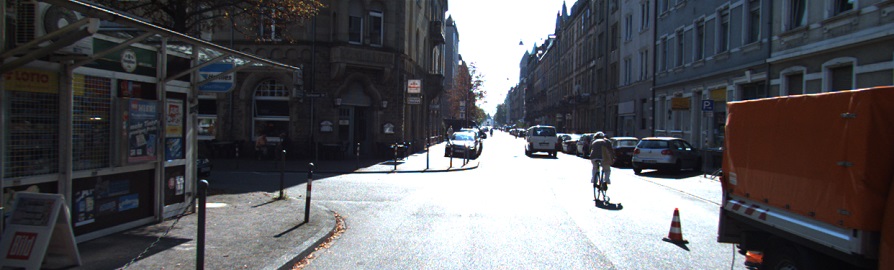}
\includegraphics[width=\columnwidth]{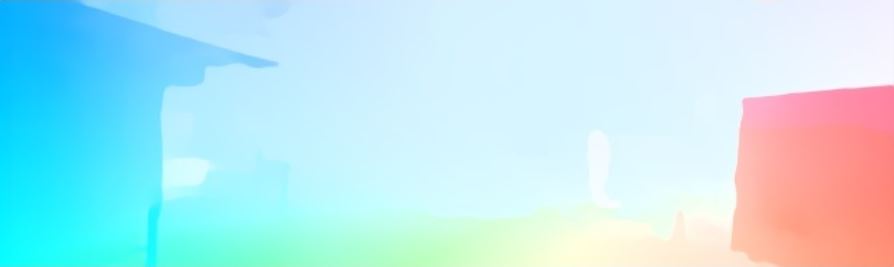}
\includegraphics[width=\columnwidth]{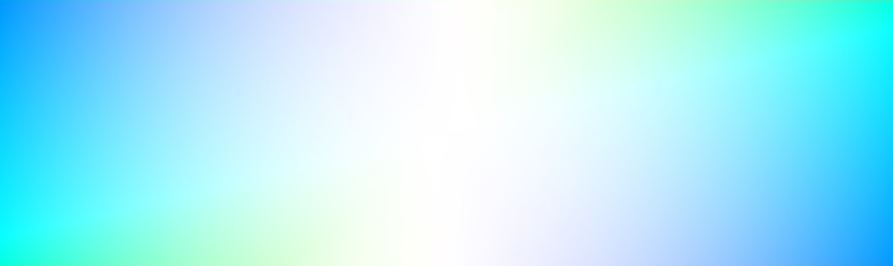}
\end{adjustbox}
\caption{Example from KITTI dataset when the camera is dominated by rotation around y-axis. From top to bottom: RGB image, Optical Flow and Vehicle Motion Tensor.}
\vspace{-0.51cm}
\label{fig:flow_samples_right}
\end{figure}

Camera rotation is mainly dominated by rotation around $y$ axis due to the steering of the car. Hence, the final VMT is mainly dominated by rotation around $y$ axis as it has larger values than rotation around $x,z$ axes as demonstrated in Fig. \ref{fig:flow_samples_right} where we show an example in KITTI dataset where the car is being steered to the right.

\subsubsection{Translation}

\begin{figure*}[!t]
\centering
\includegraphics[width=\textwidth]{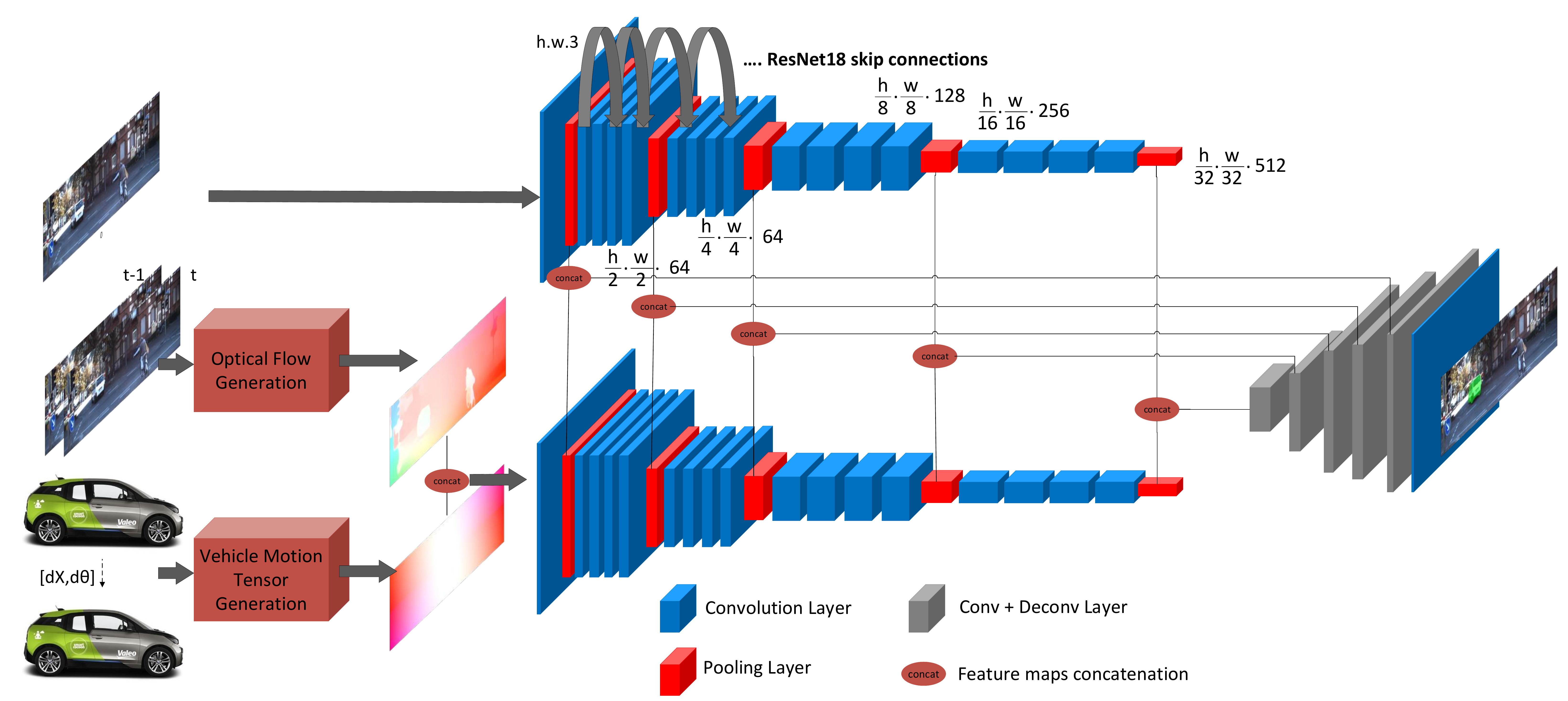}
\caption{Illustration of our VM-MODNet architecture. Vehicle Motion Tensor (VMT) is generated from vehicle's differential 3D translation $dX$ and differential 3D rotation $d\theta$. Optical flow is generated using \cite{ilg2017flownet} but this will be obtained for free in a typical automotive hardware. ResNet18 encoder is used as backbone and Deconv layers are used in the decoder.} 
\vspace{-0.51cm}
\label{fig:network_arch}
\end{figure*}

Camera translation can be described by three parameters $(T_x,T_y,T_z)$ along the three axes in 3D space. Motion field induced by pure translation is derived in  \cite{trucco1998introductory} as shown below.
\begin{align}
\vec{v_t}
=
\begin{pmatrix}
u_t \\
v_t
\end{pmatrix}
=
\dfrac{1}{Z}
\begin{pmatrix}
 T_zx -T_xf\\
 T_zy -T_yf
\end{pmatrix} 
\end{align}

where $(T_x,T_y,T_z)$ demonstrate camera translation parameters, and $Z$ indicates the scene's depth. 

Unlike $\vec{v_r}$, $\vec{v_t}$ has a scene dependent depth component. Our objective is to avoid using an additional depth measurement sensor or a more complex depth estimation network that predicts pixel-wise depth map of the scene. To simplify, we assume a constant depth virtual plane parallel to the image plane and use the motion field induced by it. The resulting motion fields are demonstrated in Fig. \ref{fig:motion_map_representations} (f,g,h). Fixed depth results in constant VMT in $x$ and $y$ directions and it doesn't add significant information to the network. However, forward motion along the $z$ direction shown in Fig. \ref{fig:motion_map_representations} (h) captures useful information across the tensor.  In this case, pixels on the right side will move to right direction (shown in shades of red) and pixels on left will move to the left direction (shown in shades of blue).




\subsection{Fusion Architectures}

We use the two-stream RGB and Optical flow mid-fusion model as the baseline architecture. We aim to keep the RGB encoder unaffected as it will be shared for other tasks in a multi-task setting. Thus we focus on different ways to fuse Vehicle Motion Tensor (VMT) with Optical Flow. Specifically, we evaluate early fusion, mid-fusion and multi-scale fusion of VMT.
We also explore weight sharing to reduce number of parameters used in the network to simplify training and to reduce model footprint. Table \ref{tab:results} summarizes results of different fusion architectures we evaluated.



\subsubsection{Early Fusion Architecture}
In early fusion, input modalities are fused before being processed by the network. In this work, we concatenate VMT and optical flow and then feed the concatenated tensor into the encoder. 
We adapt the encoder's first layer to accept an input of six channels and the corresponding weights are randomly initialized. For the rest of the encoder, ResNet18 pre-trained weights are used for initializing the training process. The output feature maps from this encoder is then concatenated with RGB encoder features and fed to the decoder. The architecture is explained in Fig. \ref{fig:network_arch}.


\subsubsection{Mid-fusion Architecture}

In mid-fusion architecture, a dedicated encoder for VMT is used and then encoder feature map is concatenated with RGB and Optical flow encoder feature maps. This increases the complexity significantly and not the preferred approach for deployment. However, we use this to understand the best possible performance. As expected, this model provides the best performance as shown in Table \ref{tab:results}.



\subsubsection{Multi-scale Fusion Architecture}

We also make use of multi-scale fusion mechanism used by CAMConvs  \cite{facil2019cam} where they fuse camera calibration information in a depth estimation network. They show that this performs better than simple concatenation. 
In our case, we perform multi-scale fusion of VMT with the optical flow encoder. We resize VMT to five different resolutions and concatenate it to the corresponding feature map while feeding to the decoder. Thus, it has slightly higher complexity than a simple early fusion. 
However, there was a slight degradation in performance compared to early fusion as reported in Table \ref{tab:results}. This is likely because optical flow and VMT are closely linked and the network is able to leverage ego-motion with simple concatenation.


\subsubsection{Weight sharing}

We aim to design efficient models targeting deployment in memory constrained automotive embedded platforms. Thus we explore weight sharing of encoders to minimize the model footprint which will be stored in persistent memory. It is more pronounced considering the large number of cameras (around 10) used in modern vehicles. Thus we perform an ablation study to understand the impact of weight sharing for both two-stream (RGB + Optical flow) and three-stream (RGB + Optical flow + VMT) architectures. Our results in Table \ref{tab:results} show that there is a significant degradation using shared weights probably due to difference in modalities of the inputs.






\section{EXPERIMENTS} \label{sec:experiments}

In this section, we provide details of the experimental setup and analysis of results obtained for different architectures.

\subsection{Dataset}
There are only a few automotive datasets that provide moving object detection (MOD) annotation. 
Cityscapes \cite{cordts2016cityscapes} has been manually labelled for MOD by \cite{vertens2017smsnet} for around 3k images. The dataset does not provide vehicle motion information which is necessary for our experiments and thus we could not use it.
KITTI \cite{Geiger2012CVPR} is the most commonly used dataset for automated driving tasks. FuseMODNet \cite{rashed2019fusemodnet} released an extended version of improved MOD annotations for 12.9k images. The dataset contains annotations for vehicles class only. In this work, we make use of this dataset and report results on the same test set used by other recent methods to enable comparison.

\subsection{Experimental Setup}
We use ResNet18 as the backbone with multiple architecture configurations explained in Section \ref{sec:methodology}. We initialize our network with ResNet18 pre-trained weights and we set the batch size to 16. The network is trained using the Ranger the Ranger (RAdam\cite{liu2019variance} + LookAhead \cite{zhang2019lookahead}) optimizer. We train all the models using weighted binary cross-entropy loss function for 60 epochs. For the early-fusion architecture, we adapt the first layer of the encoder to accept an input of 6 channels instead of 3, and we initialize the corresponding weights randomly. We use transposed convolution layers in the decoder for upsampling progressively to  the original input size.

\subsection{Results} \label{sec:results}

\begin{table}[tpb]
\centering
\caption{Quantitative comparison of different architectures. OF is Optical Flow, VMT is Vehicle Motion Tensor, + is mid-fusion architecture, x is early-fusion via concatenation of inputs, [+] is multi-scale feature map concatenation \cite{facil2019cam}, and \{\} refers to encoders with shared weights.} 
\label{tab:results}
\resizebox{0.45\textwidth}{!}{
\begin{tabular}{|l||l|l|}
\hline
\multicolumn{1}{|c||}{Architecture Type} & \multicolumn{1}{c|}{Moving IoU}       & \multicolumn{1}{c|}{mIoU}  \\ \hline
\multicolumn{3}{|c|}{\textit{RGB only architecture with shared weights}}  \\ \hline
RGB + RGB (prev)      & 40.5   & 69.85  \\ \hline

\multicolumn{3}{|c|}{\textit{RGB \& OF Fusion architectures with shared weights}}  \\ \hline
\{RGB + OF\}     & 44.7   & 72   \\ \hline
\{RGB + OF + VMT\}    & {46.6}   & {72.95}   \\ \hline

\multicolumn{3}{|c|}{\textit{RGB \& OF Fusion architectures without shared weights}}  \\ \hline
RGB + OF     & 49.3   & 74.3   \\ \hline
(RGB + OF) [+] VMT      & 51   & 75.15   \\ \hline
RGB + (OF x VMT)      & 51.4   & 75.4   \\ \hline
{RGB + \{OF  + VMT(yaw-only)\}}     & {52.9}   & 76.2   \\ \hline
RGB + \{OF  + VMT\}     & 53.6   & 76.5   \\ \hline
{RGB + OF  + VMT}     & \textbf{55.6}   & \textbf{77.6}   \\ \hline

\end{tabular}
}
\end{table}

\begin{figure*}[t!]
\captionsetup[subfigure]{labelformat=empty}
\centering

\begin{subfigure}{.33\textwidth}
    \includegraphics[width=\textwidth]{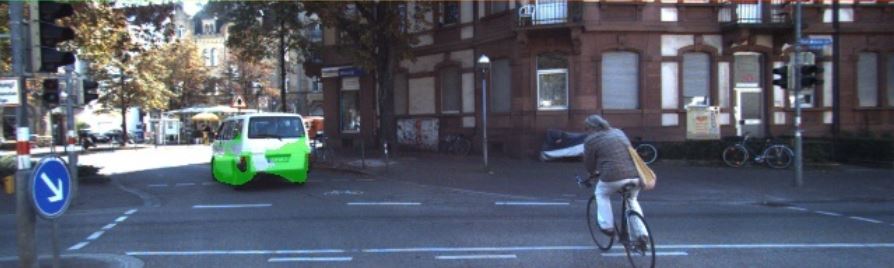}
    \vspace{-1cm}
    \caption{\textcolor{white}{(a)}}
\end{subfigure}%
\hfill
\begin{subfigure}{.33\textwidth}
    \includegraphics[width=\textwidth]{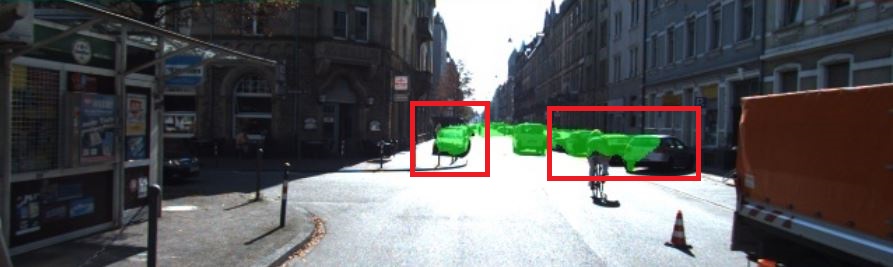}
    \vspace{-1cm}
    \caption{\textcolor{black}{(a)}}
\end{subfigure}%
\hfill
\begin{subfigure}{.33\textwidth}
    \includegraphics[width=\textwidth]{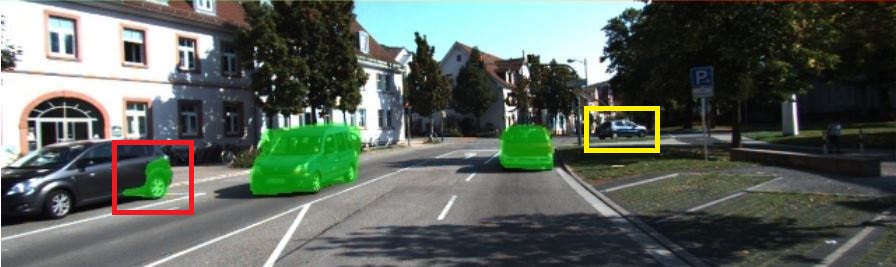}
    \vspace{-1cm}
    \caption{\textcolor{white}{(a)}}
\end{subfigure}%

\begin{subfigure}{.33\textwidth}
    \includegraphics[width=\textwidth]{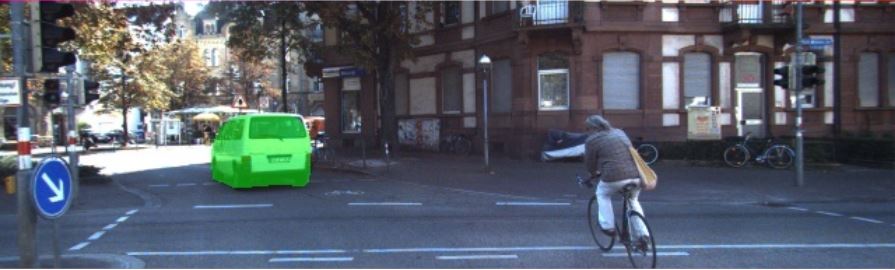}
    \vspace{-1cm}
    \caption{\textcolor{white}{(b)}}
\end{subfigure}%
\hfill
\begin{subfigure}{.33\textwidth}
    \includegraphics[width=\textwidth]{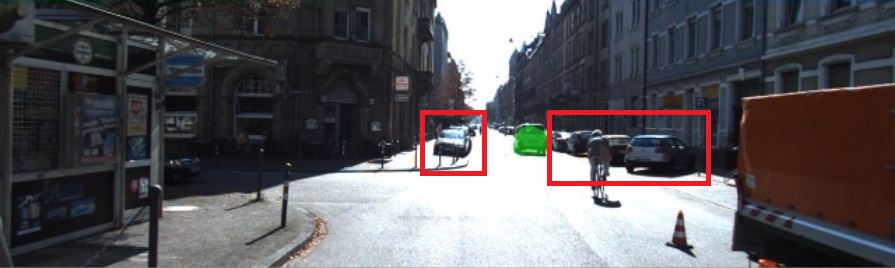}
    \vspace{-1cm}
    \caption{\textcolor{black}{(b)}}
\end{subfigure}%
\hfill
\begin{subfigure}{.33\textwidth}
    \includegraphics[width=\textwidth]{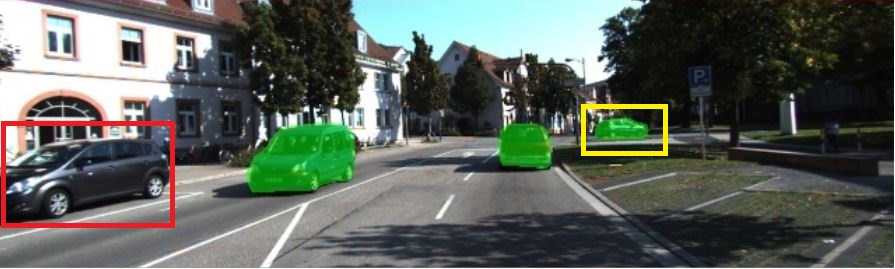}
    \vspace{-1cm}
    \caption{\textcolor{white}{(b)}}
\end{subfigure}%

\begin{subfigure}{.33\textwidth}
    \includegraphics[width=\textwidth]{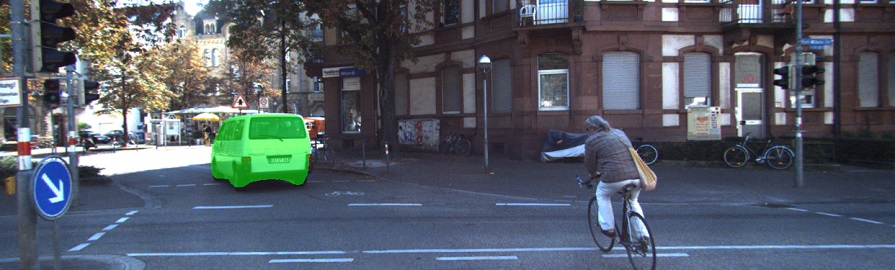}
    \vspace{-1cm}
    \caption{\textcolor{white}{(c)}}
\end{subfigure}%
\hfill
\begin{subfigure}{.33\textwidth}
    \includegraphics[width=\textwidth]{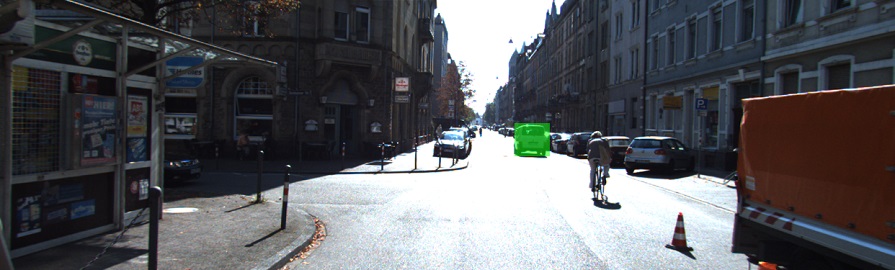}
    \vspace{-1cm}
    \caption{\textcolor{black}{(c)}}
\end{subfigure}%
\hfill
\begin{subfigure}{.33\textwidth}
    \includegraphics[width=\textwidth]{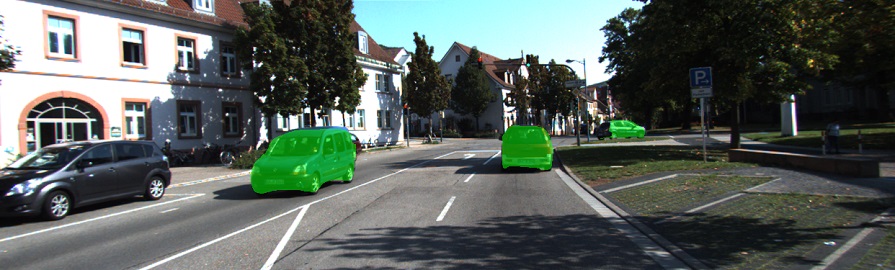}
    \vspace{-1cm}
    \caption{\textcolor{white}{(c)}}
\end{subfigure}%

    \caption{Qualitative comparison on KITTI\_MoSeg\_Extended dataset. (a) is the output of baseline RGB + Optical Flow architecture, (b) is the proposed VM-MODNet output and (c) is ground truth. Red boxes illustrates better detection of static vehicles in (b) compared to (a). Yellow box illustrates better detection of a far away moving object in (b) compared to (a).}
    \label{fig:qualitativeEval}
    \vspace{-0.51cm}
\end{figure*}

Table \ref{tab:results} demonstrates a summary of the results evaluated for different fusion architectures. The first model \textit{RGB + RGB (prev)} learns motion cues without the explicit usage of optical flow. It uses a shared encoder with two sequential RGB frames $(t,t-1)$. The shared encoder enables re-use of previous encoder without having to compute it again, thus using only one encoder in steady state. Thus it is the most efficient model. 
We obtain an accuracy of 40\% IoU for moving objects but usage of optical flow provides better performance consistent with the previous literature \cite{siam2018modnet,rashed2019fusemodnet}.

Then we report shared weight two-stream \textit{\{RGB+OF\}} and three-stream \textit{\{RGB+OF+VMT\}} architectures. \textit{\{RGB+OF\}} provides an increase of 4\% in moving IoU compared to the baseline and \textit{\{RGB+OF+VMT\}} provides an additional 2\% improvement. The same models achieve much better performance without sharing weights. The three-stream \textit{RGB+OF+VMT} achieves the best results improving moving IoU by 11\%. However, it is computationally expensive and we explore efficient two-stream versions.
We also evaluate \textit{RGB+\{OF+VMT\}} model where the weight sharing is limited only to optical flow and VMT enoders. Interestingly, its performance is closer to the model with no weight sharing than the model with shared weights. This illustrates the stronger relationship between OF and VMT.


\textit{RGB+(OF}x\textit{VMT)} is an efficient fusion model where VMT is concatenated with optical flow but it only improves performance by 2\%. However, it has the same complexity as the baseline model. We also evaluate multi-scale feature map fusion of VMT in \textit{(RGB+OF)}[\textit{+}]\textit{VMT} model but it did not provide any improvement over simple concatenation. There was a slight degradation in accuracy. In future work, we aim to explore incorporating epipolar geometric constraints as inductive bias to obtain the best performance in this efficient model.



We provide an ablation study for the effect of using only ego-vehicle rotation around y-axis (yaw or steering angle) in VMT. There are two reasons for evaluating this model \textit{RGB+{OF+VMT(yaw-only)}}. Firstly, yaw is the dominant rotation because of the steering of the vehicle. Secondly, it is available in all the vehicles using steering wheel angle measurement sensor without needing a more expensive IMU. There was only a slight degradation of 0.7\% relative to using all the rotation angles. 

\begin{table}[t]
\caption{Quantitative comparison on KITTI\_MoSeg \_Extended dataset.}
\label{tab:results_compare}
\centering
\resizebox{0.5\textwidth}{!}{
\begin{tabular}{|l||l|l||l|}
\hline
\multicolumn{1}{|c||}{Network Type} & \multicolumn{1}{c|}{\begin{tabular}[c]{@{}c@{}}Moving \\ IoU\end{tabular}}       & \multicolumn{1}{c||}{mIoU} & \multicolumn{1}{c|}{\begin{tabular}[c]{@{}c@{}}Runtime \\ (fps)\end{tabular}} \\ \hline
FuseMODNet (RGB + OF)  \cite{rashed2019fusemodnet}   & 49.36   & 74.24 & 25   \\ \hline
FuseMODNet (RGB + LiDAR)  \cite{rashed2019fusemodnet}   & 51.46   & 75.3 & 18   \\ \hline
RST-MODNet (LSTM) \cite{ramzy2019rst} & 53.3   & 76.3 & 21   \\ \hline
\hline
Ours (RGB + OF x VMT) & 51.4   & 75.4 & {\textbf{125}}   \\ \hline
Ours (RGB + OF + VMT) & \textbf{55.6}   & \textbf{77.6} & {85}   \\ \hline

\end{tabular}
}
\end{table}



Table \ref{tab:results_compare} illustrates our model performance compared to other methods. Our method achieve better performance than the other higher complexity networks which use multistage LSTM architecture \cite{ramzy2019rst} and multi-sensor model \cite{rashed2019fusemodnet} which fuses LiDAR sensor for improving motion segmentation results. Our run-time is also significantly better than these methods.

Fig. \ref{fig:qualitativeEval} shows qualitative results of the best model \textit{RGB+OF+VMT}. (a) shows the baseline results of \textit{RGB + OF} model. In some cases, we observe static cars that are incorrectly segmented as moving objects. Visually, we observe better results using VMT indicated within red boxes for static cars in (b). Furthermore, higher accuracy has been observed for moving vehicles highlighted in yellow boxes. More qualitative results can be observed in the video provide in the abstract. The proposed method provides significant improvement despite the fact that KITTI is dominated by forward motion without rotation. 


\section{CONCLUSION} \label{sec:conc}
In this paper, we proposed a vehicle motion aware moving object detection. We demonstrated significant improvements over the baseline by using vehicle motion on KITTI\_MoSeg \_Extended dataset.
Majority of this dataset comprises of the vehicle going in a straight line with a standard urban driving velocity and we expect larger improvements on a more diverse ego-motion dataset. We perform a comparative study on different types of network architectures and fusion mechanisms to find the best model which achieves state-of-the-art results. Given the emergence of low-cost IMU sensors for commercial deployment, we hope that our work encourages further research in using vehicle motion for other tasks including tracking and depth estimation.







\bibliographystyle{ieee}
\bibliography{references/egbib}

\end{document}